\documentclass[10pt,letterpaper,twocolumn]{article}

\usepackage{cvpr}
\usepackage{times}
\usepackage[accsupp]{axessibility}

\usepackage{color,xcolor}
\usepackage{epsfig}
\usepackage{graphicx}

\usepackage{adjustbox}
\usepackage{array}
\usepackage{booktabs}
\usepackage{colortbl}
\usepackage{float,wrapfig}
\usepackage{hhline}
\usepackage{multirow}
\usepackage{subcaption} 
\usepackage[toc,page]{appendix}

\usepackage{amsmath,amsfonts,amsthm,amssymb}
\usepackage{bm}
\usepackage{nicefrac}
\usepackage{microtype}
\usepackage{inconsolata}
\usepackage{pifont}

\usepackage{changepage}
\usepackage{extramarks}
\usepackage{fancyhdr}
\usepackage{lastpage}
\usepackage{setspace}
\usepackage{soul}
\usepackage{xspace}
\usepackage{indentfirst}
\usepackage{paralist}

\usepackage[pagebackref=true,breaklinks=true,letterpaper=true,colorlinks,citecolor=citecolor,bookmarks=false]{hyperref}
\usepackage{url}

\usepackage{algorithm, algorithmic}
\usepackage{enumerate}
\usepackage{lipsum}
\newcolumntype{L}[1]{>{\raggedright\let\newline\\\arraybackslash\hspace{0pt}}m{#1}}
\newcolumntype{C}[1]{>{\centering\let\newline\\\arraybackslash\hspace{0pt}}m{#1}}
\newcolumntype{R}[1]{>{\raggedleft\let\newline\\\arraybackslash\hspace{0pt}}m{#1}}

\newcommand{\fig}[1]{Figure~\ref{#1}}

\newcommand{\tab}[1]{Table~\ref{#1}}

\newcommand{\ignorethis}[1]{}

\makeatletter
\DeclareRobustCommand\onedot{\futurelet\@let@token\@onedot}
\def\@onedot{\ifx\@let@token.\else.\null\fi\xspace}

\def\eg{\emph{e.g}\onedot} 
\def\ie{\emph{i.e}\onedot} 
 
 \def\vs{\emph{vs}\onedot}
 
\def\etal{\emph{et al}\onedot}
\makeatother

\definecolor{citecolor}{HTML}{0071bc}
\definecolor{mydarkblue}{rgb}{0,0.08,1}
\definecolor{mydarkgreen}{rgb}{0.02,0.6,0.02}
\definecolor{mydarkred}{rgb}{0.8,0.02,0.02}
\definecolor{mydarkorange}{rgb}{0.40,0.2,0.02}
\definecolor{mypurple}{RGB}{111,0,255}
\definecolor{myred}{rgb}{1.0,0.0,0.0}
\definecolor{mygold}{rgb}{0.75,0.6,0.12}
\definecolor{mydarkgray}{rgb}{0.66, 0.66, 0.66}

\definecolor{darkblue}{rgb}{0,0.08,1}
\definecolor{darkgreen}{rgb}{0.02,0.6,0.02}
\definecolor{darkred}{rgb}{0.8,0.02,0.02}
\definecolor{darkorange}{rgb}{0.40,0.2,0.02}
\definecolor{darkpurple}{RGB}{111,0,255}

\newcommand{\myparagraph}[1]{\vspace{-8.5pt}\paragraph{#1}}

\def\model{SparseViT\xspace}

\def\cvprPaperID{9744}
\def\confName{CVPR}
\def\confYear{2023}

\begin{document}

\title{\model: Revisiting Activation Sparsity for \\ Efficient High-Resolution Vision Transformer}

\author{Xuanyao Chen\textsuperscript{1,2,$*$} \hspace{5mm} Zhijian Liu\textsuperscript{4,$*$} \hspace{5mm} Haotian Tang\textsuperscript{4} \hspace{5mm} Li Yi\textsuperscript{1,3} \hspace{5mm} Hang Zhao\textsuperscript{1,3} \hspace{5mm} Song Han\textsuperscript{4} \\
\textsuperscript{1}Shanghai Qi Zhi Institute \hspace{5mm}
\textsuperscript{2}Fudan University \hspace{5mm}
\textsuperscript{3}Tsinghua University \hspace{5mm}
\textsuperscript{4}MIT \\\\
\url{https://sparsevit.mit.edu}
}

\maketitle

\footnotetext{$*$ indicates equal contributions (listed in alphabetical order).}

\begin{abstract}

High-resolution images enable neural networks to learn richer visual representations. However, this improved performance comes at the cost of growing computational complexity, hindering their usage in latency-sensitive applications. As not all pixels are equal, skipping computations for less-important regions offers a simple and effective measure to reduce the computation. This, however, is hard to be translated into actual speedup for CNNs since it breaks the regularity of the dense convolution workload. In this paper, we introduce \textbf{\model} that revisits activation sparsity for recent window-based vision transformers (ViTs). As window attentions are naturally batched over blocks, actual speedup with window activation pruning becomes possible: \ie, $\sim$50\% latency reduction with 60\% sparsity. Different layers should be assigned with different pruning ratios due to their diverse sensitivities and computational costs. We introduce sparsity-aware adaptation and apply the evolutionary search to efficiently find the optimal layerwise sparsity configuration within the vast search space. SparseViT achieves speedups of \textbf{1.5$\times$}, \textbf{1.4$\times$}, and \textbf{1.3$\times$} compared to its dense counterpart in monocular 3D object detection, 2D instance segmentation, and 2D semantic segmentation, respectively, with negligible to no loss of accuracy.

\end{abstract}
\section{Introduction}

\begin{figure}[t]
    \centering
    \includegraphics[width=\linewidth]{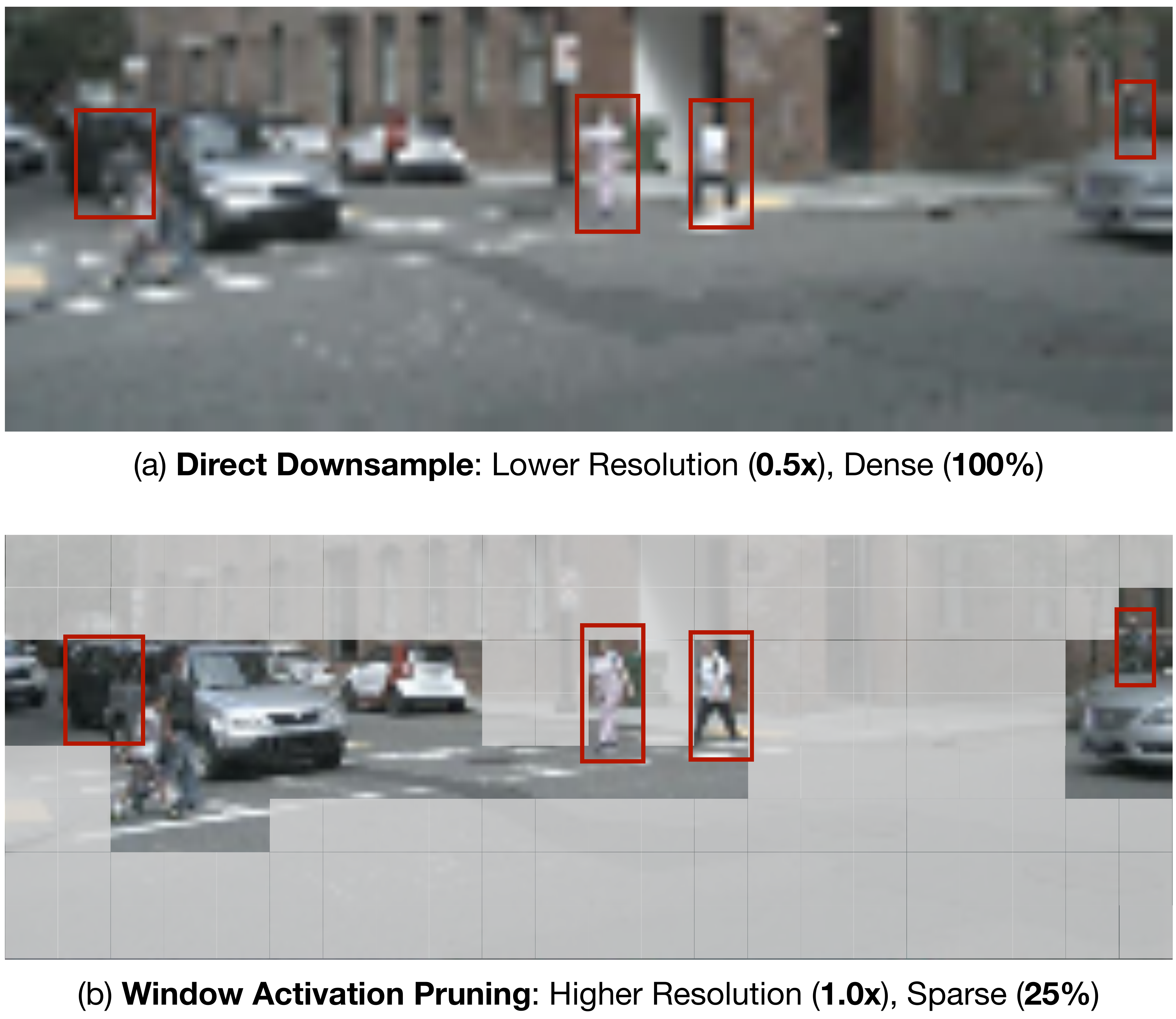}
    \caption{\textit{Sparse, high-resolution} features are far more informative than \textit{dense, low-resolution} ones. Compared with direct downsampling, activation pruning can retain important details at a higher resolution, which is essential for most image recognition tasks.}
    \label{fig:teaser}
    \vspace{-8pt}
\end{figure}

With the advancement of image sensors, high-resolution images become more and more accessible: \eg, recent mobile phones are able to capture 100-megapixel photos. The increased image resolution offers great details and enables neural network models to learn richer visual representations and achieve better recognition quality. This, however, comes at the cost of linearly-growing computational complexity, making them less deployable for resource-constrained applications (\eg, mobile vision, autonomous driving).

The simplest solution to address this challenge is to downsample the image to a lower resolution. However, this will drop the fine details captured from the high-resolution sensor. What a waste! The missing information will bottleneck the model's performance upper bound, especially for small object detection and dense prediction tasks. For instance, the detection accuracy of a monocular 3D object detector will degrade by more than 5\% in mAP by reducing the height and width by 1.6$\times$\footnote{BEVDet~\cite{huang2021bevdet} (with Swin Transformer~\cite{liu2021swin} as backbone) achieves 31.2 mAP with 256$\times$704 resolution and 25.90 mAP with 160$\times$440 resolution.}. Such a large gap cannot be easily recovered by scaling the model capacity up.

Dropping details uniformly at all positions is clearly sub-optimal as not all pixels are equally informative (\fig{fig:teaser}\textcolor{red}{a}). Within an image, the pixels that contain detailed object features are more important than the background pixels. Motivated by this, a very natural idea is to skip computations for less-important regions (\ie, activation pruning). However, activation sparsity cannot be easily translated into the actual speedup on general-purpose hardware (\eg, GPU) for CNNs. This is because sparse activation will introduce randomly distributed and unbalanced zeros during computing and cause computing unit under-utilization~\cite{xu2022falcon}. Even with dedicated system support~\cite{ren2018sbnet}, a high sparsity is typically required to realize speedup, which will hurt the model's accuracy.

Recently, 2D vision transformers (ViTs) have achieved tremendous progress. Among them, Swin Transformer~\cite{liu2021swin} is a representative work that generalizes well across different visual perception tasks (such as image classification, object detection, and semantic segmentation). Our paper revisits the activation sparsity in the context of window-based ViTs. Different from convolutions, window attentions are naturally batched over windows, making real speedup possible with window-level activation pruning. We re-implement the other layers in the model (\ie, \texttt{FFN}s and \texttt{LN}s) to also execute at the window level. As a result, we are able to achieve around 50\% latency reduction with 60\% window activation sparsity.

Within a neural network, different layers have different impacts on efficiency and accuracy, which advocates for a non-uniform layerwise sparsity configuration: \eg, we may prune layers with larger computation and lower sensitivity more, while pruning layers with smaller computation and higher sensitivity less. To this end, we make use of the evolutionary search to explore the best per-layer pruning ratios under a resource constraint. We also propose \emph{sparsity-aware adaptation} by randomly pruning a different subset of the activations at each iteration. This effectively adapts the model to activation sparsity and avoids the expensive re-training of every candidate within the large search space. Our SparseViT achieves speedups of \textbf{1.5$\times$}, \textbf{1.4$\times$}, and \textbf{1.3$\times$} compared to its dense counterpart in monocular 3D object detection, 2D instance segmentation, and 2D semantic segmentation, respectively, with negligible to no loss of accuracy.
\section{Related Work}

\paragraph{Vision Transformers.}

Transformers~\cite{vaswani2017attention} have revolutionized natural language processing (NLP) and are now the backbone of many large language models (LLMs)~\cite{devlin2018bert}. Inspired by their success, researchers have explored the use of transformers in a range of visual recognition tasks~\cite{khan2021transformers}. ViT~\cite{kolesnikov2021vit} was the first work in this direction, demonstrating that an image can be divided into 16$\times$16 words and processed using multi-head self-attention. DeiT~\cite{touvron2021training} improves on ViT's data efficiency. T2T-ViT~\cite{yuan2021t2tvit}, Pyramid ViT~\cite{wang2021pyramid,wang2021pvtv2}, and CrossFormer~\cite{wang2021crossformer} introduce hierarchical modeling capability to ViTs. Later, Swin Transformer~\cite{liu2021swin,liu2022swinv2} applies self-attention to non-overlapping windows and proposes window shifting to enable cross-window feature communication. There have also been extensive studies on task-specific ViTs, such as ViTDet~\cite{li2022exploring} for object detection, and SETR~\cite{zheng2021rethinking} and SegFormer~\cite{xie2021segformer} for semantic segmentation.

\myparagraph{Model Compression.}

As the computational cost of neural networks continues to escalate, researchers are actively investigating techniques for model compression and acceleration~\cite{han2016deep,he2018amc}. One approach is to design more efficient neural network architectures, either manually~\cite{iandola2016squeezenet,howard2017mobilenets,sandler2018mobilenetv2,ma2018shufflenet,zhang2018shufflenet} or using automated search~\cite{cai2019proxylessnas,cai2020once,guo2020single,zoph2017neural,liu2021pvnas,tang2020searching}. These methods are able to achieve comparable performance to ResNet~\cite{he2016deep} with lower computational cost and latency. Another active direction is neural network pruning, which involves removing redundant weights at different levels of granularity, ranging from unstructured~\cite{han2015learning,han2016deep} to structured~\cite{liu2017learning,he2017channel}. Although unstructured pruning can achieve higher compression ratios, the lower computational cost may not easily translate into measured speedup on general-purpose hardware and requires specialized hardware support. Low-bit weight and activation quantization is another approach that has been explored to reduce redundancy and speed up inference~\cite{jacob2018quantization,he2016deep,wang2019haq,wang2020hardware}.

\myparagraph{Activation Pruning.}

Activation pruning differs from static weight pruning as it is dynamic and input-dependent. While existing activation pruning methods typically focus on reducing memory cost during training~\cite{raihan2020sparse,liu2018dynamic,liu2022spatial}, few of them aim to improve inference latency as activation sparsity does not always lead to speedup on hardware. To overcome this, researchers have explored adding system support for activation sparsity~\cite{ren2018sbnet,tang2022torchsparse,yan2018second}. However, these libraries often require extensive engineering efforts and high sparsity rates to achieve measurable speedup over dense convolutions.

\myparagraph{Efficient ViTs.}

Several recent works have explored different approaches to improve the efficiency of ViTs. For instance, MobileViT~\cite{mehta2022mobilevit} combines CNN and ViT by replacing local processing in convolutions with global processing using transformers. MobileFormer~\cite{chen2021mobile} parallelizes MobileNet and Transformer with a two-way bridge for feature fusing, while NASViT~\cite{gong2022nasvit} leverages neural architecture search to find efficient ViT architectures. Other works have focused on token pruning for ViTs~\cite{yin2021adavit,kong2021spvit,pan2021ia,rao2021dynamicvit,wang2021spatten,kim2021learned,tang2022patch,chen2021chasing}. However, these approaches mainly focus on token-level pruning, which is finer-grained than window pruning.
\section{\model}

\begin{figure*}[t]
    \centering
    \includegraphics[width=\linewidth]{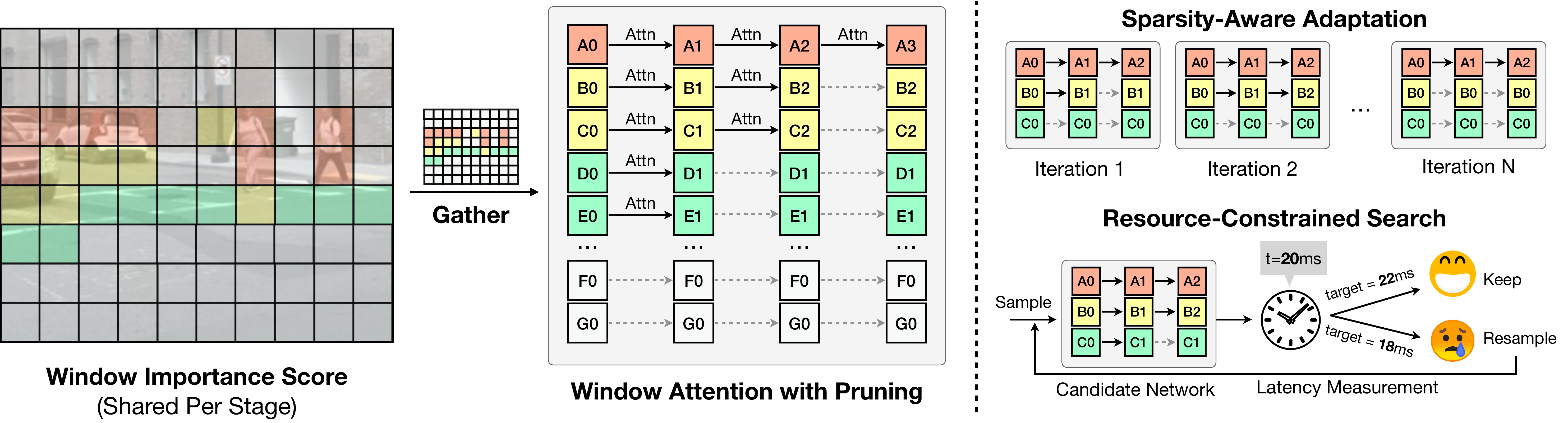}
    \caption{Overview of \model. \textbf{(Left)} \model first computes the L2 norm of each window activation as its importance score. After that, it first gathers the features from the windows with the highest importance scores, then runs self-attention on selected windows, and finally scatter the results back. \textbf{(Right upper)} \model leverages sparsity-aware adaptation that samples a different layerwise activation sparsity at each training iteration to accommodate the activation sparsity. \textbf{(Right lower)} \model makes use of evoluationary search to explore the best layerwise sparsity configuration given a latency constraint.}
    \label{fig:overview}
    \vspace{-8pt}
\end{figure*}

In this section, we first briefly revisit Swin Transformer and modify its implementation so that all layers are applied to windows. We then introduce how to incorporate the window activation sparsity into the model. Finally, we describe an efficient algorithm (based on sparsity-aware adaptation and evolutionary search) to find the layerwise sparsity ratio.

\subsection{Swin Transformer Revisited}
Swin Transformer~\cite{liu2021swin} applies multi-head self-attention (\texttt{MHSA}) to extract local features within non-overlapping image windows (\eg, 7$\times$7). The transformer design follows a standard approach, involving layer normalization (\texttt{LN}), \texttt{MHSA}, and a feed-forward layer (\texttt{FFN}) applied to each window. Swin Transformer uses a shifted window approach that alternates between two different window partitioning configurations to introduce cross-window connections efficiently.

\myparagraph{Window-Wise Execution.}

The original implementation of Swin Transformer applies \texttt{MHSA} at the window level, while \texttt{FFN} and \texttt{LN} are applied to the entire feature map. This mismatch between the two operations requires additional structuring before and after each \texttt{MHSA}, making window pruning more complicated as the sparsity mask must be mapped from the window level to the feature map. To simplify this process, we modify the execution of \texttt{FFN} and \texttt{LN} to be window-wise as well. This means that all operations will be applied at the window level, making the mapping of the sparsity mask very easy. In practice, this modification incurs a minimal accuracy drop of less than 0.1\% due to padding, even without re-training. By making the execution of all operations window-wise, our method simplifies the pruning process.

\subsection{Window Activation Pruning}
\label{sect:method:window_activation_pruning}

Not all windows are equally important. In this paper, we define the importance of each window as its L2 activation magnitude. This is much simpler than other learning-based measures since it introduces smaller computational overhead and is fairly effective in practice.

Given an activation sparsity ratio (which will be detailed in the next section), we first gather those windows with the highest importance scores, then apply \texttt{MHSA}, \texttt{FFN} and \texttt{LN} only on these selected windows, and finally scatter outputs back. \fig{fig:overview} shows the workflow of window activation pruning. To mitigate information loss due to coarse-grained window pruning, we simply duplicate the features of the unselected windows. This approach incurs no additional computation, yet proves highly effective in preserving information, which is critical for dense prediction tasks such as object detection and semantic segmentation.

\begin{table*}[t] 
\setlength{\tabcolsep}{3.5pt}
\small\centering
\begin{tabular}{lccccccccccc}
\toprule
Backbone & Resolution & Width & \#MACs (G) & Latency (ms) & mAP$_\uparrow$ & mATE$_\downarrow$ & mASE$_\downarrow$ & mAOE$_\downarrow$ & mAVE$_\downarrow$ & mAAE$_\downarrow$ & NDS$_\uparrow$ \\
\midrule
Swin-T & 256$\times$704 & 1$\times$ & 140.8 & 36.4 & 31.2 & 69.1 & 27.2 & 52.3 & 90.9 & 24.7 & 39.2\\
\textbf{\model} (Ours) & 288$\times$792 & 1$\times$ & 113.9 & 34.5 & \textbf{32.0} & 72.8 & 27.2 & 53.8 & 79.4 & 25.7 & \textbf{40.1} \\ 

\midrule
Swin-T (R224) & 224$\times$616 & 1$\times$ & 78.5 & 23.0 & 29.9 & 71.8 & 27.4 & 60.9 & 79.0 & 26.0 & 38.4\\
Swin-T (W0.6$\times$) & 256$\times$704 & 0.6$\times$ & 56.0 & 22.6 & 29.9 & 69.9 & 27.5 & 59.9 & 81.4 & 25.8 & 38.5 \\
\textbf{\model} (Ours) & 256$\times$704 & 1$\times$ & 78.4 & 23.8 & \textbf{31.2} & 70.9 & 27.5 & 58.7 & 83.1 & 27.2 & \textbf{38.9} \\ 

\midrule
Swin-T (R192) & 192$\times$528 & 1$\times$ & 67.1 & 18.7 & 28.7 & 74.3 & 27.9 & 59.5 & 76.7 & 27.8 & 37.7\\
Swin-T (W0.4$\times$) & 256$\times$704 & 0.4$\times$ & 20.4 & 17.6 & 27.6 & 74.2 & 27.9 & 63.4 & 91.0 & 26.2 & 35.5\\
\textbf{\model} (Ours) & 256$\times$704 & 1$\times$ & 58.6 & 18.7 & \textbf{30.0} & 72.0 & 27.5 & 59.7 & 81.7 & 26.6 & \textbf{38.3} \\ 
\bottomrule
\end{tabular}
\caption{Results of monocular 3D object detection on nuScenes.}
\label{tab:results:nuscenes}
\end{table*}  

\myparagraph{Shared Scoring.}

Unlike conventional weight pruning, importance scores are input-dependent and need to be computed during inference, which can introduce significant overhead to the latency. To mitigate this, we compute the window importance score only once per stage and reuse it across all the blocks within the stage, amortizing the overhead. This also ensures that the window ordering remains consistent within a stage. We simplify the gathering operation using slicing, which does not require any feature copying.

\subsection{Mixed-Sparsity Configuration Search}

Using a uniform sparsity level throughout a model may not be the best strategy because different layers have varying impacts on both accuracy and efficiency. For example, early layers typically require more computation due to their larger feature map sizes, while later layers are more amenable to pruning as they are closer to the output. Thus, it is more beneficial to apply more pruning to layers with lower sensitivity and higher costs. However, manually exploring layerwise sparsity can be a time-consuming and error-prone task. To overcome this challenge, we propose a workflow that efficiently searches for the optimal mixed-sparsity pruning.

\myparagraph{Search Space.}

We first design the search space for mixed-sparsity activation pruning. For each Swin block, we allow the sparsity ratio to be chosen from $\{0\%, 10\%, \ldots, 80\%\}$. Note that each Swin block contains two \texttt{MHSAs}, one with shifted window and one without. We will assign them with the same sparsity ratio. Also, we enforce the sparsity ratio to be non-descending within each stage. This ensures that a pruned window will not engage in the computation again.

\myparagraph{Sparsity-Aware Adaptation.}

To identify the best mixed-sparsity configuration for a model, it is crucial to evaluate its accuracy under different sparsity settings. However, directly assessing the original model's accuracy with sparsity might produce unreliable results (see Section~\ref{sect:analysis}). On the other hand, retraining the model with all possible sparsity configurations before evaluating its accuracy is impractical due to the significant time and computational costs involved. We therefore propose \textit{sparsity-aware adaptation} as a more practical solution to address this challenge. Our approach involves adapting the original model, which was trained with only dense activations, by randomly sampling layerwise activation sparsity and updating the model accordingly at each iteration. After adaptation, we can obtain a more accurate estimate of the performance of different sparsity configurations without the need for full retraining. This enables us to efficiently and effectively evaluate different mixed-sparsity configurations and identify the optimal one for the model. Notably, our approach differs from super network training (used in NAS) as we only randomly sample activations, without changing the number of parameters.

\myparagraph{Resource-Constrained Search.}

With an accurate estimate of the model's performance through sparsity-aware adaptation, we can proceed to search for the best sparsity configurations within specified resource constraints. In this paper, we consider two types of resource constraints: hardware-agnostic theoretical computational cost, represented by the number of multiply-accumulate operations (\#MACs), and hardware-dependent measured latency. To perform the sparsity search, we adopt the evolutionary algorithm~\cite{guo2020single}. We first initialize the population with $n$ randomly sampled networks within the search space and using rejection sampling (\ie, repeated resampling until satisfaction) to ensure every candidate meets the specified resource constraint. In each generation, we evaluate all individuals in the population and select the top $k$ candidates with the highest accuracy. We then generate the population for the next generation through $(n/2)$ mutations and $(n/2)$ crossovers using rejection sampling to satisfy the hard resource constraints. We repeat this process to obtain the best configuration from the population in the last generation.

\myparagraph{Finetuning with Optimal Sparsity.}

The resulting model from our resource-constrained search has been trained under a variety of sparsity configurations during the adaptation stage. To further optimize its performance, we finetune the model with the fixed sparsity configurations identified in the search process until convergence.

\section{Experiments}

\begin{table*}[t]
\setlength{\tabcolsep}{6pt}
\small\centering
\begin{tabular}{lcccccccccc}
\toprule
Backbone & Resolution & Width & \#MACs (G) & Latency (ms) & $\text{AP}^\text{bbox}$ & $\text{AP}^\text{bbox}_\text{50}$ & $\text{AP}^\text{bbox}_\text{75}$ & $\text{AP}^\text{mask}$ & $\text{AP}^\text{mask}_\text{50}$ & $\text{AP}^\text{mask}_\text{75}$  \\
\midrule
Swin-T & 640$\times$640 & 1$\times$ & 161.8 & 46.6 & 42.0 & 63.3 & 45.7 & 38.3 & 60.3 & 40.9\\
\midrule
Swin-T (R576) & 576$\times$576 & 1$\times$ & 149.5 & 41.3 & 41.0 & 62.1 & 44.9 & 37.2 & 59.0 & 39.6\\
Swin-T (W0.9$\times$) & 640$\times$640 & 0.9$\times$ & 122.3 & 41.8 & 40.4 & 61.9 & 43.8 & 37.1 & 58.9 & 39.8\\
\textbf{\model} (Ours) & 672$\times$672 & 1$\times$ & 139.5 & 41.3 & \textbf{42.4} & 63.3 & 46.4 & \textbf{38.5} & 60.3 & 41.3\\ 
\midrule
Swin-T (R544) & 544$\times$544 & 1$\times$ & 119.8 & 34.8 & 40.5 & 61.2 & 43.8 & 36.8 & 58.2 & 39.1\\
Swin-T (W0.8$\times$) & 640$\times$640 & 0.8$\times$ & 90.5 & 35.9 & 39.4 & 60.7 & 42.8 & 36.4 & 57.9 & 38.8 \\
\textbf{\model} (Ours) & 672$\times$672 & 1$\times$ & 116.5 & 34.1 & \textbf{41.6} & 62.5 & 45.5 & \textbf{37.7} & 59.4 & 40.2\\ 
\midrule
Swin-T (R512) & 512$\times$512 & 1$\times$ & 117.5 & 32.9 & 39.6 & 60.1 & 43.4 & 36.0 & 57.0 & 38.2\\
Swin-T (W0.6$\times$) & 640$\times$640 & 0.6$\times$ & 63.4 & 31.7 & 38.7 & 60.2 & 41.6 & 35.7 & 57.0 & 38.0 \\
\textbf{\model} (Ours) & 672$\times$672 & 1$\times$ & 105.9 & 32.9 & \textbf{41.3} & 62.2 & 44.9 & \textbf{37.4} & 59.1 & 39.7 \\ 
\bottomrule
\end{tabular}
\caption{Results of 2D instance segmentation on COCO.}
\label{tab:results:coco}
\vspace{-8pt}
\end{table*}

In this section, we evaluate our method on three diverse tasks, including monocular 3D object detection, 2D instance segmentation, and 2D semantic segmentation.

\myparagraph{Latency Measurement.}

We report the latency of the backbone in all our results as our method is only applied to the backbone. The latency measurements are obtained using a single NVIDIA RTX A6000 GPU. To ensure accurate measurements, we perform 500 inference steps as a warm-up and subsequently measure the latency for another 500 steps. To minimize the variance, we report the average latency of the middle 250 measurements out of the 500.

\subsection{Main Results}

\subsubsection{Monocular 3D Object Detection}

\paragraph{Dataset and Metrics.}

We use nuScenes~\cite{nuscenes} as the benchmark dataset for monocular 3D object detection, which includes 1k scenes with multi-modal inputs from six surrounding cameras, one LiDAR, and five radars. We only employ camera inputs in our experiments. We report official metrics, including mean average precision (mAP), average translation error (ATE), average scale error (ASE), average orientation error (AOE), average velocity error (AVE), and average attribute error (AAE). We also report the nuScenes detection score (NDS), which is a weighted average of the six metrics.

\myparagraph{Model and Baselines.}

We use BEVDet~\cite{huang2021bevdet} as the base model for monocular 3D object detection. It adopts Swin-T~\cite{liu2021swin} as the baseline and employs FPN~\cite{lin2017focal} to fuse information from multi-scale features. Following BEVDet~\cite{huang2021bevdet}, we resize the input images to 256$\times$704 and train the model for 20 epochs. We compare our SparseViT against two common model compression strategies: reducing resolution and width. For reduced resolution, we re-train the model with different resolutions. For reduced width, we uniformly scale down the model to 0.4$\times$ and 0.6$\times$, then pre-train it on ImageNet~\cite{deng2009imagenet} and finally finetune it on nuScenes~\cite{nuscenes}.

\myparagraph{Compared with Reduced Resolution.}

The accuracy of monocular 3D object detection is highly influenced by resolution scaling. With fine-grained features in higher-resolution images, our SparseViT outperforms the baseline with smaller resolutions, with comparable or faster latency. The results in Table \ref{tab:results:nuscenes} show that \model achieves the same accuracy as Swin-T with \textbf{1.8$\times$} lower \#MACs and \textbf{1.5$\times$} faster inference latency. Furthermore, when compared to the baseline with 192$\times$528 resolution, \model achieves 30.0 mAP and 38.3 NDS at 50\% latency budget of the full Swin-T backbone, which is \textbf{1.3} mAP and 0.6 NDS better, respectively.

\myparagraph{Comparison with Reduced Width.}

Reducing the model's width lowers \#MACs. However, this decrease in computation cost might not necessarily translate into a measured speedup due to low device utilization. SparseViT outperforms the baseline with 0.6$\times$ width by 1.3 mAP and the one with 0.4$\times$ width by 2.4 mAP at similar latency. This indicates that activation pruning is more effective than model pruning in latency-oriented compression.

\begin{figure*}[t]
    \centering
     \includegraphics[width=\linewidth]{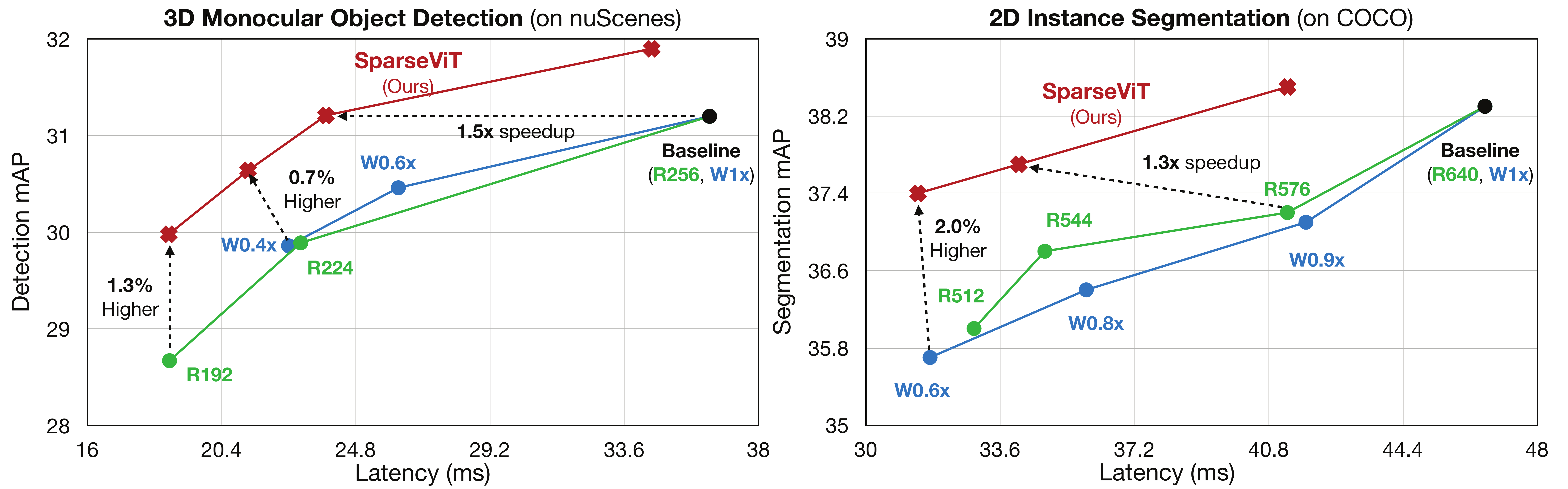}
    \caption{SparseViT delivers a significantly better accuracy-efficiency trade-off than the baselines with reduced resolutions and widths on monocular 3D object detection (\textbf{left}) and 2D instance segmentation (\textbf{right}).}
    \label{fig:results:main}
    \vspace{-8pt}
\end{figure*}

\subsubsection{2D Instance Segmentation}

\paragraph{Dataset and Metrics.}

We use COCO~\cite{lin2014microsoft} as our benchmark dataset for 2D instance segmentation, which contains 118k/5k training/validation images. We report the box/mask average precision (AP) over 50\% to 95\% IoU thresholds.

\myparagraph{Model and Baselines.}

We use Mask R-CNN~\cite{he2017mask} as the base model. The model uses Swin-T~\cite{liu2021swin} as its backbone. We adopt 640$\times$640 as the default input resolution and train the model for 36 epochs. We compare our SparseViT against baselines with reduced resolutions and widths. For reduced resolution, we train the model using random scaling augmentation~\cite{he2017mask,liu2021swin} and evaluate the model under different resolutions. For reduced width, we uniformly scale down the model to 0.6$\times$, 0.8$\times$ and 0.9$\times$, pre-train it on ImageNet~\cite{deng2009imagenet} and finetune it on COCO~\cite{lin2014microsoft}.

\myparagraph{Comparison with Reduced Resolution.}

As in \tab{tab:results:coco}, SparseViT consistently outperforms the baseline with less computation across various input resolutions from 512$\times$512 to 640$\times$640. Our key insight is that starting with a high resolution of 672$\times$672 and aggressively pruning the activation is more efficient than directly scaling down the input resolution. This observation aligns with the visualizations in \fig{fig:teaser}, where fine-grained details become indistinguishable under low resolution. Despite using a higher resolution, SparseViT achieves 1.2$\times$ smaller \#MACs than the baseline while delivering 0.4\% higher AP\textsuperscript{bbox} and 0.2\% higher AP\textsuperscript{mask}. With similar accuracy, SparseViT has \textbf{1.4$\times$} lower \#MACs, resulting in a perfect \textbf{1.4$\times$} speedup. This is because our SparseViT performs \textit{window-level} activation pruning, which is equivalent to reducing the batch size in \texttt{MHSA} computation and is easy to accelerate on hardware. Similarly, to match the accuracy of the baseline with 90\% resolution, SparseViT is \textbf{1.3$\times$} faster and consumes \textbf{1.4$\times$} less computation. Remarkably, despite using 30\% larger resolution (\ie, 1.7$\times$ larger \#MACs to begin with!), SparseViT is more efficient than the baseline at 512$\times$512 resolution, while providing significantly better accuracy (\textbf{+1.7} AP\textsuperscript{bbox} and \textbf{+1.4} AP\textsuperscript{mask}).

\myparagraph{Comparison with Reduced Width.}

In Table \ref{tab:results:coco}, we also compare \model with the baseline with reduced channel width. Although reducing channel width leads to a significant reduction in \#MACs, we do not observe a proportional increase in speed. For example, the baseline with 0.6$\times$ channel width on 640$\times$640 inputs consumes only 63.4G MACs, yet it runs slightly slower than \model on 672$\times$672 inputs with 105.9G MACs (which is actually 1.7$\times$ heavier!). GPUs prefer wide and shallow models to fully utilize computation resources. Pruning channels will decrease device utilization and is not as effective as reducing the number of windows (which is equivalent to directly reducing the batch size in \texttt{MHSA} computation) for model acceleration.

\subsubsection{2D Semantic Segmentation}

\begin{table}[t] 
\setlength{\tabcolsep}{7pt}
\small\centering
\begin{tabular}{lccc}
\toprule
 Backbone & Resolution & Latency (ms) & mIoU \\
\midrule
Swin-L & 1024$\times$2048 & 329.5 & 83.3\\
\midrule
Swin-L (R896) & \,\,\,896$\times$1792 & 256.5 & 82.8 \\
\textbf{SparseViT} (Ours) & 1024$\times$2048 & 250.6 & \textbf{83.2} \\
\bottomrule
\end{tabular}
\caption{Results of 2D semantic segmentation on Cityscapes.}
\label{tab:mask2former}
\vspace{-8pt}
\end{table}

\paragraph{Dataset and Metrics.}

Our benchmark dataset for 2D semantic segmentation is Cityscapes~\cite{cordts2016cityscapes}, which consists of over 5k high-quality, fully annotated images with pixel-level semantic labels for 30 object classes, including road, building, car, and pedestrian. We report mean intersection-over-union (mIoU) as the primary evaluation metric on this dataset.

\myparagraph{Model and Baselines.}

We use Mask2Former~\cite{cheng2021mask2former} as our base model, which uses Swin-L~\cite{liu2021swin} as its backbone. We train the model for 90k iterations with an input resolution of 1024$\times$2048. Here, we only compare to the baseline with reduced resolution.

\myparagraph{Results.}

Based on \tab{tab:mask2former}, SparseViT model attains comparable segmentation accuracy to the baseline while delivering a speedup of 1.3$\times$. In contrast, reducing the resolution results in a more substantial decrease in accuracy. By utilizing spatial redundancy, SparseViT delivers competitive results while being more efficient than direct downsampling.

\subsection{Analysis}
\label{sect:analysis}

\begin{figure*}[!t]
    \centering
    \includegraphics[width=\linewidth]{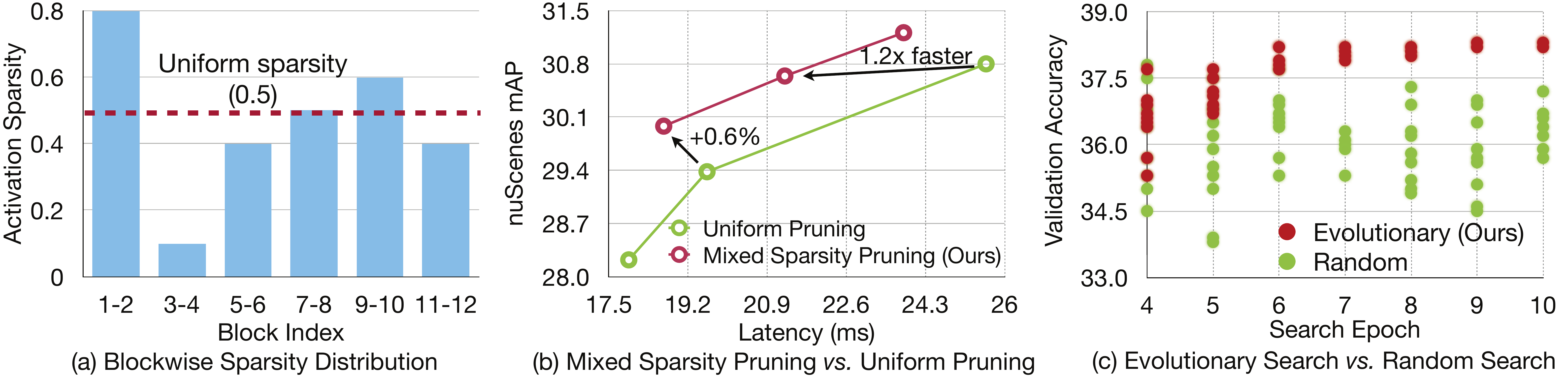}
    \caption{Non-uniform sparsity is better than uniform sparsity (a, b). Evolutionary search is more sample-efficient than random search (c).}
    \label{fig:ablations:search}
    \vspace{-8pt}
\end{figure*}

In this section, we present analyses to validate the effectiveness of our design choices. All studies are conducted on monocular 3D object detection (on nuScenes), except the evolutionary search one in \fig{fig:ablations:search}(c), which is conducted on 2D instance segmentation (on COCO).

\myparagraph{Window pruning is more effective than token pruning.}

\tab{tab:dynvit} demonstrates that SparseViT achieves a low computational cost and latency without any loss of accuracy, whereas DynamicViT~\cite{rao2021dynamicvit}, a learnable token pruning method, experiences a substantial decrease in accuracy of 0.4 mAP with only a minor reduction in computational cost. These findings offer valuable insights into the comparative performance of these pruning methods. Furthermore, it is worth noting that token pruning requires more fine-grained \textit{token-level} gathering, which has inferior memory access locality and tends to be slower on GPUs, unlike window pruning in our SparseViT that only necessitates \textit{window-level} gathering.

\begin{table}[!h]
\setlength{\tabcolsep}{7pt}
\small\centering
\begin{tabular}{cccc}
\toprule
 & \#MACs (G) & Latency (ms)  & mAP Drop \\
\midrule
DynamicViT & 98.6 & 33.3 & -0.1 \\ 
DynamicViT & 91.7 & 32.6 & -0.4 \\
\midrule
SparseViT & \textbf{78.4} & \textbf{23.8} & \textbf{0.0} \\
\bottomrule
\end{tabular}
\caption{Window pruning (SparseViT) is more efficient and effective than learnable token pruning (DynamicViT).}
\label{tab:dynvit}
\vspace{-8pt}
\end{table}

\myparagraph{Pruning from higher resolution matters.} 

A key design insight in SparseViT is that it is more advantageous to start with a \textit{high-resolution} input and \textit{prune more}, rather than beginning with a \textit{low-resolution} input and \textit{pruning less}. While counterintuitive, starting with a high-resolution input allows us to retain fine-grained information in the image. The abundance of uninformative background windows provides us with ample room for activation pruning. Quantitatively, as in \tab{tab:prune_from_higher}, starting from the highest resolution (\ie, 256$\times$704) produces the best accuracy under the same latency constraint.

\begin{table}[!h]
\setlength{\tabcolsep}{8pt}
\small\centering
\begin{tabular}{cccc}
\toprule
Input Resolution & \#MACs (G) & Latency (ms)  & mAP  \\
\midrule
192$\times$528 & 67.1 & \textbf{18.7} & 28.7 \\ 
224$\times$616 & 64.9 & 19.1 & 29.7 \\
256$\times$704 & \textbf{58.6} & \textbf{18.7} & \textbf{30.0} \\
\bottomrule
\end{tabular}
\caption{Starting from a high-resolution input and pruning more is better than starting from a low-resolution input and pruning less.}
\label{tab:prune_from_higher}
\vspace{-8pt}
\end{table}

\myparagraph{Mixed-sparsity pruning is better than uniform pruning.}

In \fig{fig:ablations:search}(a), we show the pruning strategy used by SparseViT to achieve 50\% overall sparsity. Unlike uniform sparsity ratios applied to all layers, SparseViT favors non-uniform sparsity ratios for different layers based on their proximity to the input. Specifically, the smaller window sizes in the first and second blocks allow for more aggressive pruning, while larger window sizes in later layers result in less aggressive pruning. This non-uniform sparsity selection leads to better accuracy, as in \fig{fig:ablations:search}(b). Compared to uniform pruning, SparseViT achieves similar accuracy but is up to 1.2$\times$ faster. Alternatively, when compared at similar speeds, SparseViT achieves 0.6\% higher accuracy than uniform pruning.

\begin{figure*}[t]
    \captionsetup[subfigure]{labelformat=empty}
  \subfloat[]{\includegraphics[width=0.12\textwidth]{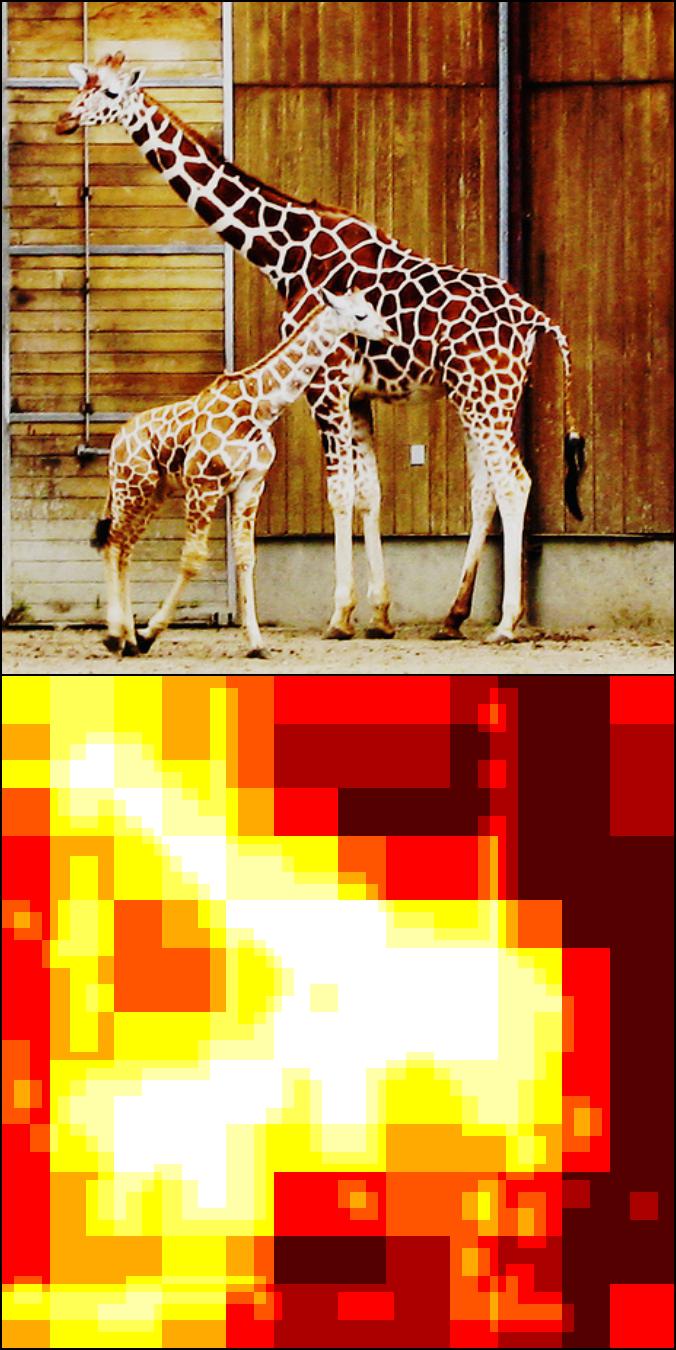}}
 \hfill 	
  \subfloat[]{\includegraphics[width=0.12\textwidth]{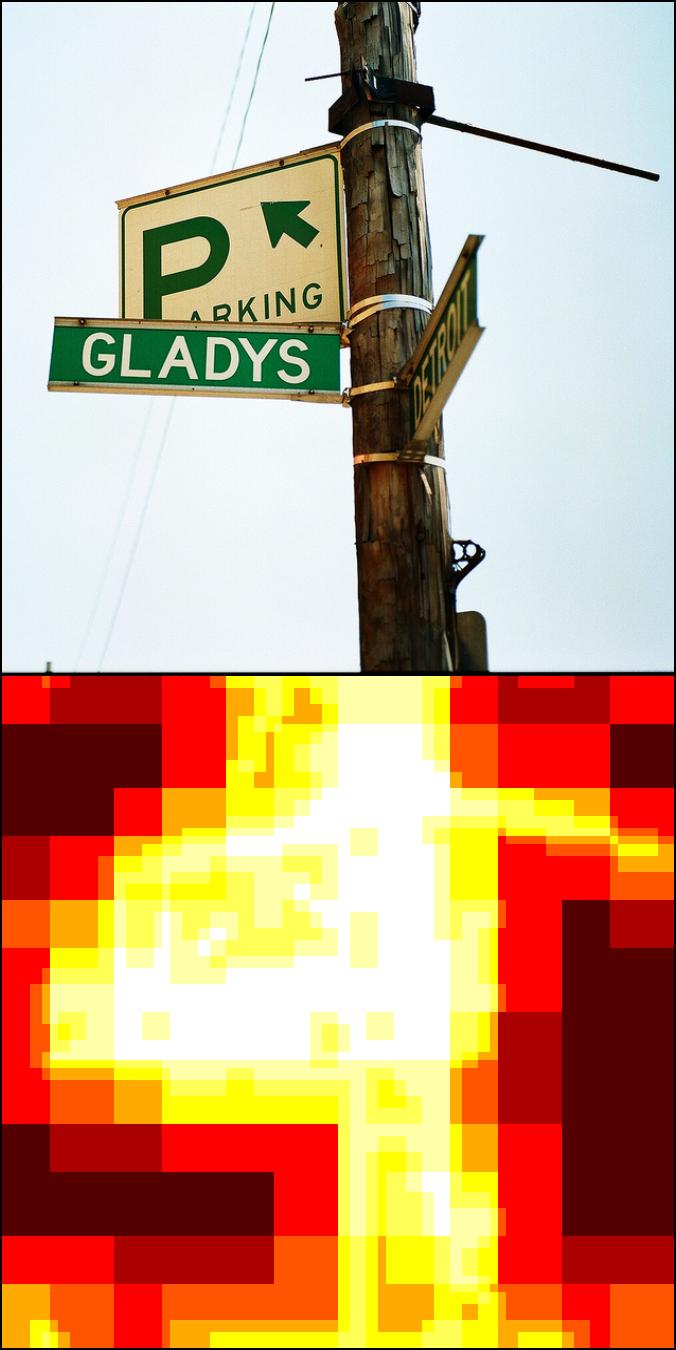}}
 \hfill	
  \subfloat[]{\includegraphics[width=0.12\textwidth]{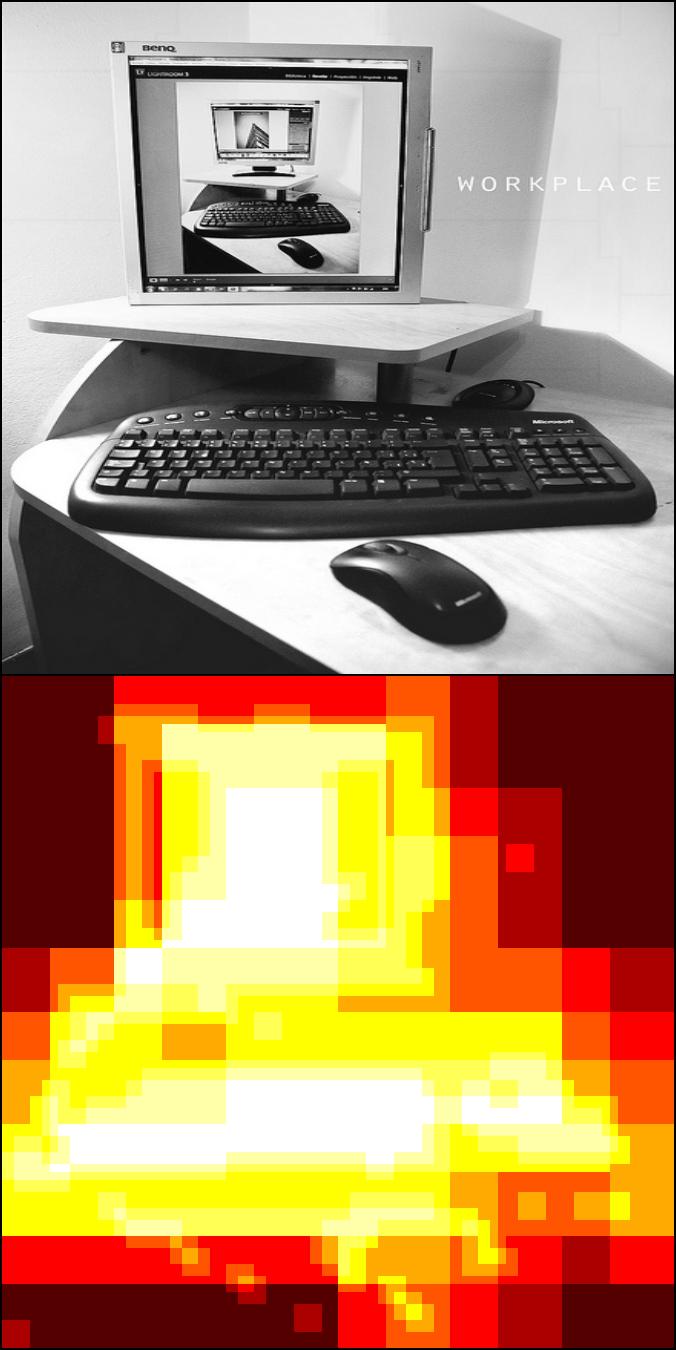}}
  \hfill
  \subfloat[]{\includegraphics[width=0.12\textwidth]{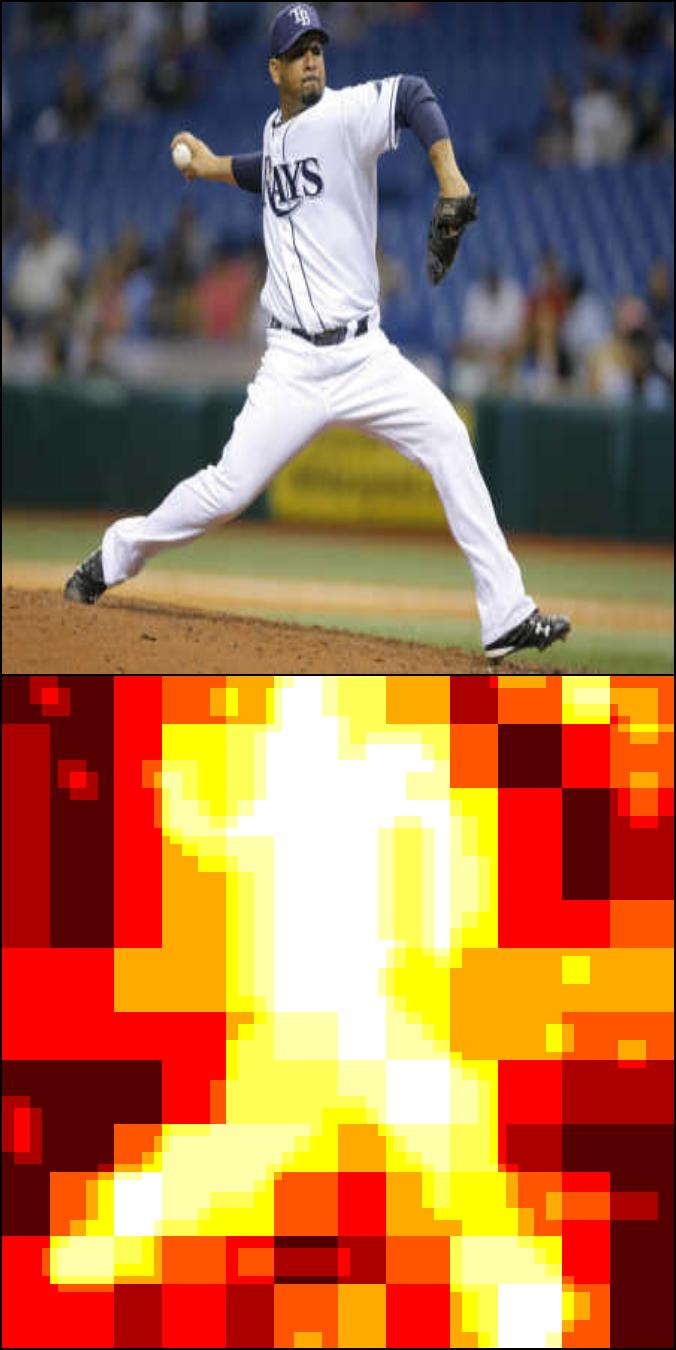}}
  \hfill
  \subfloat[]{\includegraphics[width=0.12\textwidth]{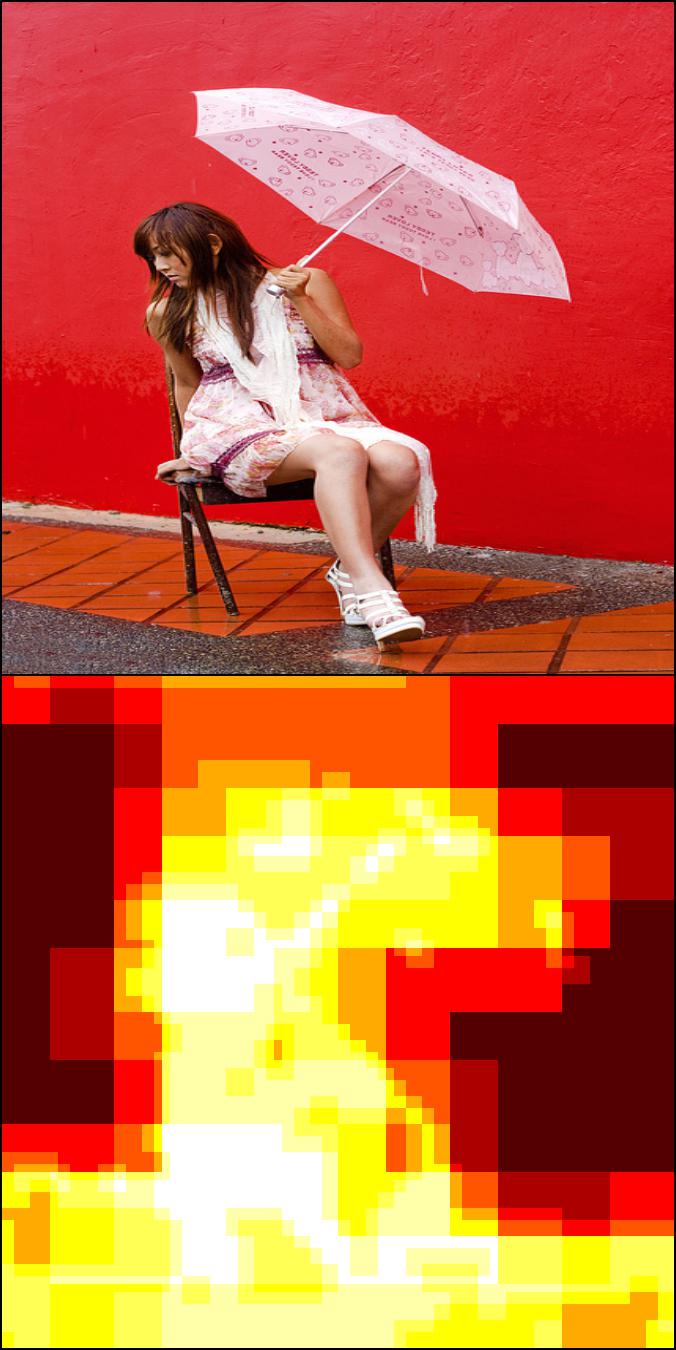}}
  \hfill
  \subfloat[]{\includegraphics[width=0.12\textwidth]{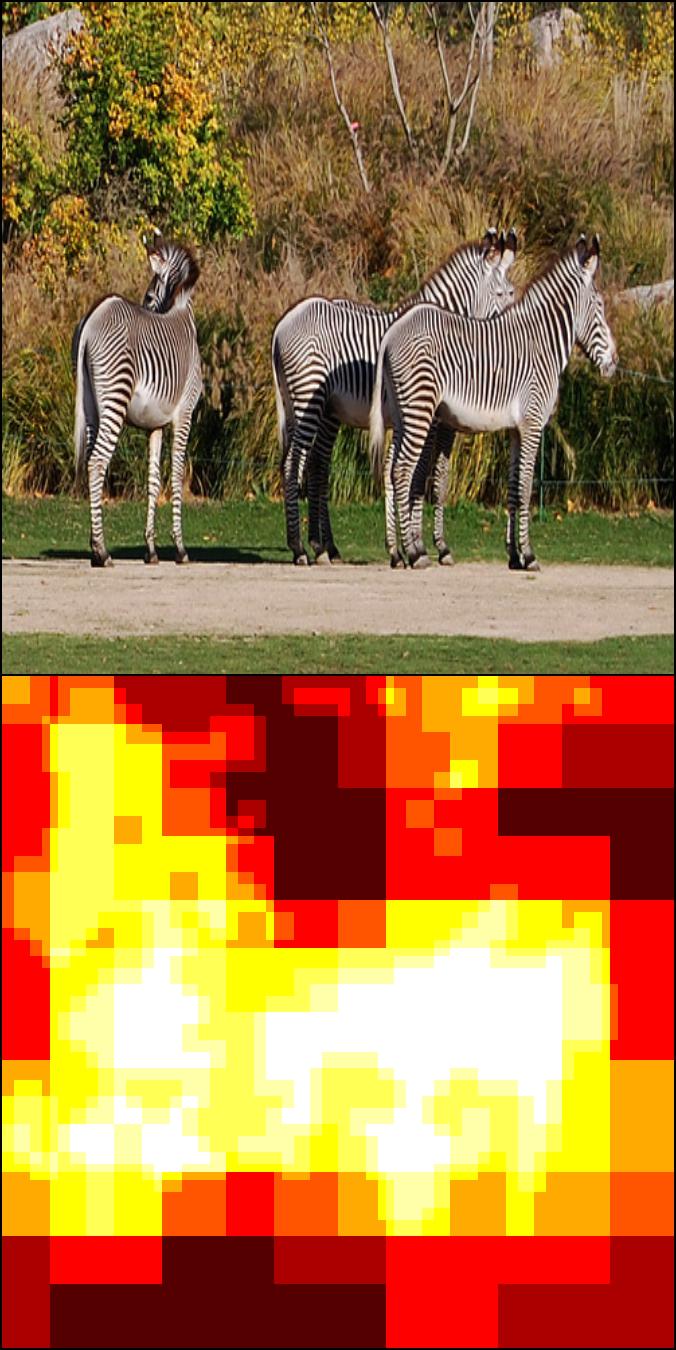}}
  \hfill
  \subfloat[]{\includegraphics[width=0.12\textwidth]{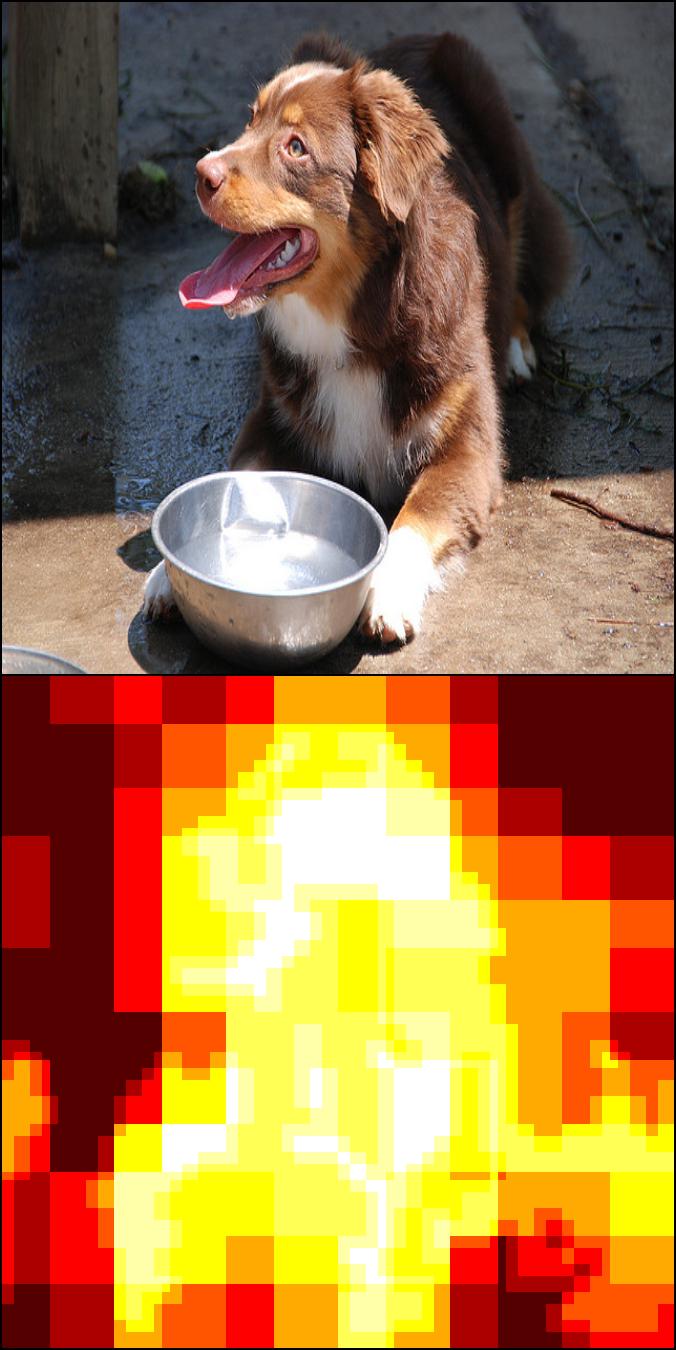}}
  \hfill
  \subfloat[]{\includegraphics[width=0.12\textwidth]{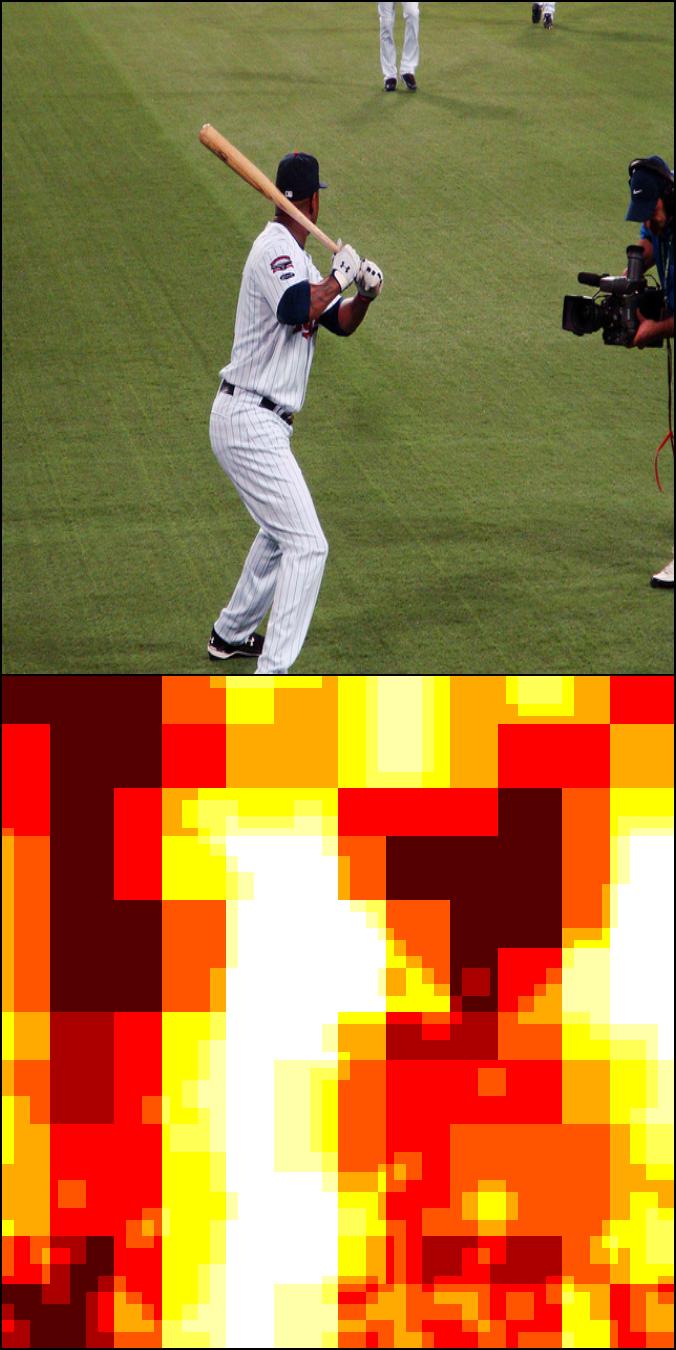}}

  \vspace{-5pt}
 
  \subfloat[]{\includegraphics[width=0.16\textwidth]{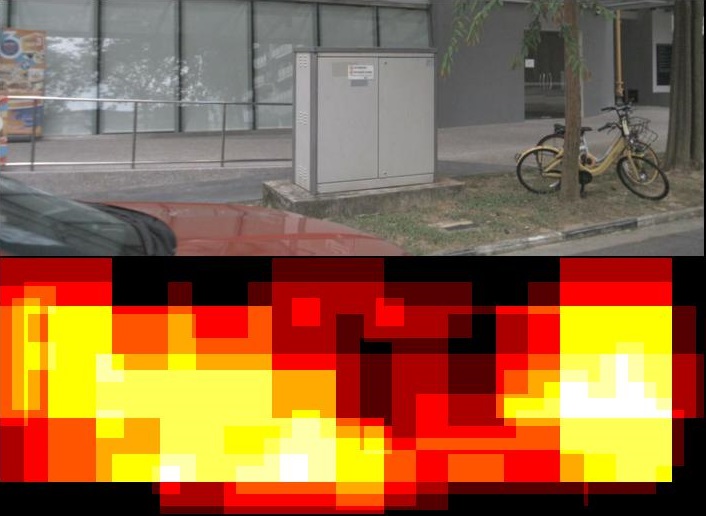}}
 \hfill 	
  \subfloat[]{\includegraphics[width=0.16\textwidth]{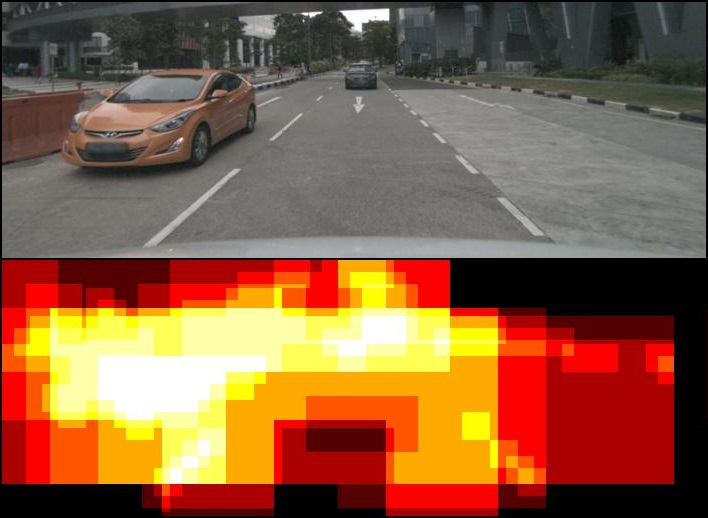}}
 \hfill	
  \subfloat[]{\includegraphics[width=0.16\textwidth]{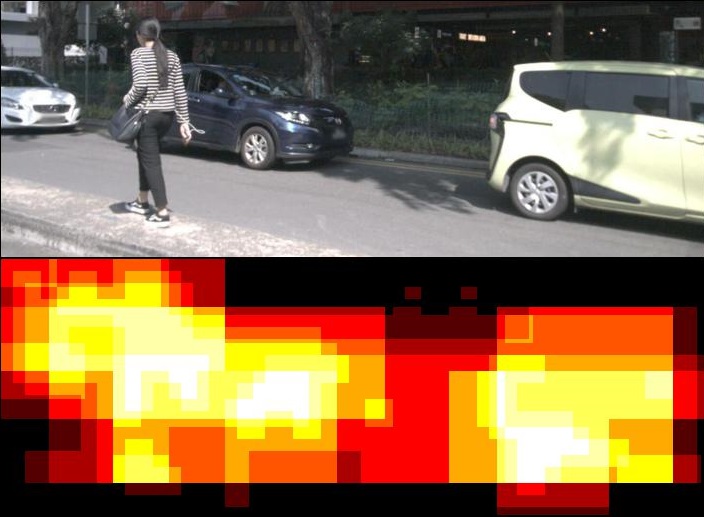}}
  \hfill	
  \subfloat[]{\includegraphics[width=0.16\textwidth]{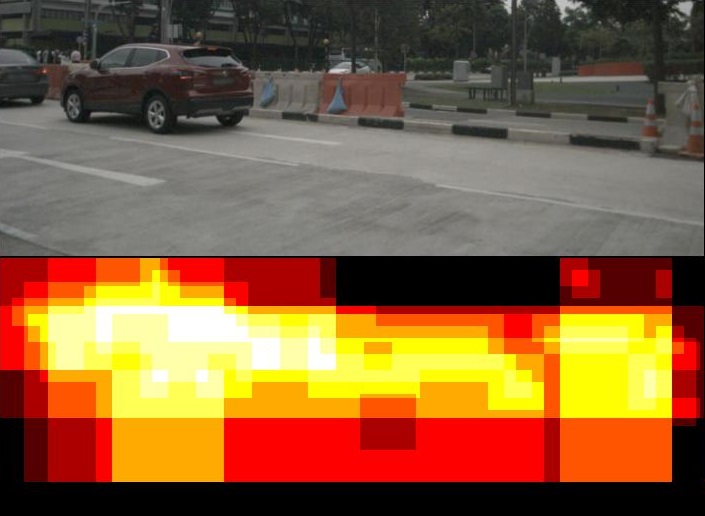}}
  \hfill	
  \subfloat[]{\includegraphics[width=0.16\textwidth]{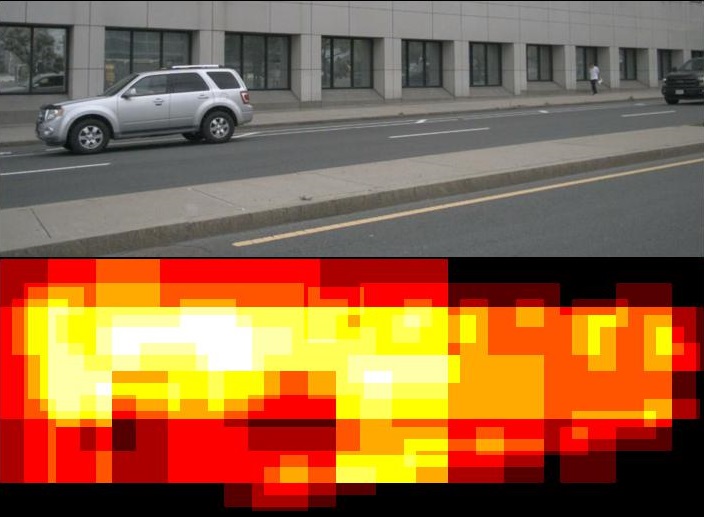}}
  \hfill	
  \subfloat[]{\includegraphics[width=0.16\textwidth]{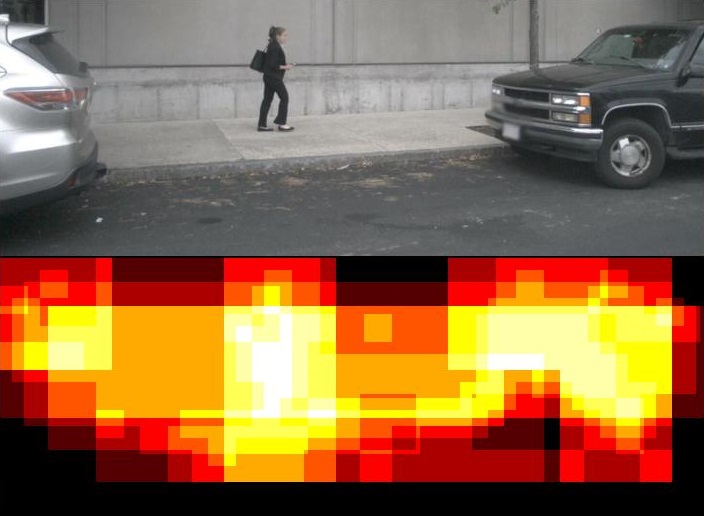}}
\vspace{-15pt}
\caption{SparseViT effectively prunes irrelevant background windows while retaining informative foreground windows. Each window's color corresponds to the number of layers it is executed. Brighter colors indicate that the model has executed the window in more layers.}
\label{fig:visualization}
\vspace{-8pt}
\end{figure*}

\myparagraph{Evolutionary search is better than random search.}

We demonstrate the efficacy of evolutionary search in selecting the sparsity ratios. Figure \ref{fig:ablations:search}(c) compares the results obtained by evolutionary search and random search, by visualizing the validation mAP of the best-performing models found in the last seven epochs. The accuracy of the models discovered by evolutionary search converges to a high value of 37.5 after the sixth epoch, whereas the models found by random search still exhibit high variance until the final epoch.

\myparagraph{Sparsity-aware adaptation offers a better proxy.}

Conventional pruning methods~\cite{han2015learning,han2016deep} typically scan different sparsity ratios at each layer, zero out corresponding weights, and evaluate accuracy directly on a holdout validation set. While this approach works well for weight pruning in CNNs, it falls short when generalizing to activation pruning in ViTs. For example, as in \tab{tab:sensitivity}, the naive accuracy sensitivity analysis approach asserts that pruning away 50\% of the windows in stage 2 is much better than doing so in stage 1 (\ie, 30.93 mAP \vs 28.31 mAP). However, if both models are finetuned, the accuracy difference is almost zero. In contrast, sparsity-aware adaptation provides a much better accuracy proxy (\ie, 30.83 mAP \vs 30.63 mAP), which can serve as a better feedback signal for evolutionary search.

\begin{table}[!h] 
\setlength{\tabcolsep}{9pt}
\small\centering
\begin{tabular}{cccc}
\toprule
SAA & Sparsity Ratios & mAP w/o FT & mAP w/ FT  \\
\midrule
 & (0.5, 0, 0, 0) & 28.31 & 31.40 \\
 & (0, 0.5, 0, 0) & 30.93 & 31.37 \\
 \midrule
\multirow{2}{*}{$\checkmark$} & (0.5, 0, 0, 0) & 30.83 & 31.45 \\
 & (0, 0.5, 0, 0) & 30.63 & 31.44 \\
\bottomrule
\end{tabular}
\caption{Sparsity-aware adaptation (SAA) improves the correlation between the accuracy before and after finetuning (FT).}
\label{tab:sensitivity}
\vspace{-8pt}
\end{table}

\myparagraph{Sparsity-aware adaptation improves the accuracy.}

In \tab{tab:super_ft}, we explore the impact of sparsity-aware adaptation on model convergence during the finetuning stage. Using the same window sparsity ratios, we compare three approaches: \textbf{(a)} training from scratch, \textbf{(b)} finetuning from a pre-trained Swin-T, and \textbf{(c)} finetuning from a pre-trained Swin-T with sparsity-aware adaptation (SAA). Our results indicate that approach \textbf{(c)} achieves the best performance. We speculate that sparsity-aware adaptation could serve as a form of implicit distillation during training. Models with higher window sparsity can benefit from being co-trained with models with lower window sparsity (and therefore higher \#MACs and higher capacity), resulting in improved overall performance.

\begin{table}[h] 
\small\centering
\begin{tabular}{lc}
\toprule
& mAP \\
\midrule
\textbf{(a)} Training from scratch & 29.67 \\
\textbf{(b)} Finetuning from a pre-trained Swin-T & 30.83 \\
\textbf{(c)} Finetuning from a pre-trained Swin-T with SAA & \textbf{31.21} \\
\bottomrule
\end{tabular}
\caption{Sparsity-aware adaptation improves the convergence.}
\label{tab:super_ft}
\vspace{-8pt}
\end{table}

\myparagraph{SparseViT keeps important foreground windows.}

In \fig{fig:visualization}, we visualize the window pruning strategy discovered by SparseViT, where the color represents the number of layers each window is executed. Notably, on the first row, SparseViT automatically learns the contour of the objects, as demonstrated in the third and fourth figures, where the computer and sportsman are respectively outlined. Furthermore, on the second row, the windows corresponding to foreground objects are not pruned away. Despite being a small object, the pedestrian in the last figure is retained throughout the entire execution, illustrating the effectiveness of SparseViT.

\myparagraph{L2 magnitude-based scoring is simple and effective.}

\tab{tab:scoring} demonstrates that the L2 magnitude-based scoring is simple and effective, outperforming the learnable window scoring that utilizes MLP and Gumbel-Softmax for predicting window scores. We also include the regularization loss on pruning ratios, following Rao~\etal~\cite{rao2021dynamicvit}, to restrict the proportion of preserved windows to a predetermined value in the learnable window scoring baseline. However, the added complexity of the learnable scoring results in higher latency and \#MACs compared to the L2 magnitude-based scoring. Achieving an optimal balance between pruning regularization loss and detection loss is not an easy task, as evidenced by a 0.4 mAP drop observed in the learnable scoring method.

\begin{table}[h] 
\small\centering
\begin{tabular}{cccc}
\toprule
Scoring & \#MACs (G) & Latency (ms) & mAP Drop \\
\midrule
Learnable & 89.6 & 33.1 &-0.6 \\
L2 Magnitude & \textbf{78.4} & \textbf{23.8} & \textbf{0.0} \\
\bottomrule
\end{tabular}
\caption{L2 magnitude-based scoring is simple and effective, achieving a better accuracy-efficiency trade-off than learnable scoring.}
\label{tab:scoring}
\vspace{-8pt}
\end{table}

\myparagraph{Shared scoring is better than independent scoring.}

We compare our proposed shared scoring strategy with the independent scoring. Despite being an approximation, sharing window scores per stage, as in Table \ref{tab:shared_stage}, does not negatively impact the performance. This strategy amortizes the cost of score calculation, which allows for more effective computation and results in a better accuracy-latency trade-off.

\begin{table}[h] 
\small\centering
\setlength{\tabcolsep}{5pt}
\begin{tabular}{cccccc}
\toprule
Scoring & Sparsity & \#MACs (G) & Latency (ms)  & mAP \\
\midrule
Independent  & 0.5 & 70.68 & 23.3 & 30.08 \\
Shared  & 0.5 & 70.68 & \textbf{23.0} & \textbf{30.17} \\ 
\bottomrule
\end{tabular}
\caption{Shared scoring per stage reduces the cost of score calculation, leaving room for effective computation and offering a better accuracy-latency trade-off than independent scoring per block.}
\label{tab:shared_stage}
\vspace{-8pt}
\end{table}

\section{Conclusion}

Although activation pruning is a very powerful technique for preserving high-resolution information, it does not offer actual speedup for CNNs. In this paper, we revisit activation sparsity for recent window-based ViTs and propose a novel approach to leverage it. We introduce sparsity-aware adaptation and employ evolutionary search to efficiently find the optimal layerwise sparsity configuration. As a result, \model achieves 1.5$\times$, 1.4$\times$, and 1.3$\times$ measured speedups in monocular 3D object detection, 2D instance segmentation, and 2D semantic segmentation, respectively, with minimal to no loss in accuracy. We hope that our work inspires future research to explore the use of activation pruning for achieving better efficiency while retaining high-resolution information.
\myparagraph{Acknowledgement.}

This work was supported by National Science Foundation (NSF), MIT-IBM Watson AI Lab, MIT AI Hardware Program, Amazon-MIT Science Hub, NVIDIA Academic Partnership Award, and Hyundai. Zhijian Liu was partially supported by Qualcomm Innovation Fellowship.

{
\small
\bibliographystyle{ieee}
\bibliography{reference}
}

\end{document}